\documentclass[10pt, conference]{IEEEtran}
\IEEEoverridecommandlockouts

\usepackage[left=0.75in, right=0.75in, top=0.95in, bottom=1in]{geometry}

\usepackage{cite}
\usepackage{amsmath,amssymb,amsfonts}
\usepackage{graphicx}
\usepackage{textcomp}
\usepackage{xcolor}
\usepackage{nccmath}
\usepackage[hidelinks]{hyperref}
\def\BibTeX{{\rm B\kern-.05em{\sc i\kern-.025em b}\kern-.08em
    T\kern-.1667em\lower.7ex\hbox{E}\kern-.125emX}}

\IEEEoverridecommandlockouts
\IEEEpubid{\makebox[\columnwidth]{IEEE CogMI 2021 978-1-6654-4875-8/21/\$31.00~\copyright2021 IEEE \hfill} \hspace{\columnsep}\makebox[\columnwidth]{}}

\begin{document}

\fontsize{9}{11}\selectfont

\title{\vspace{0.2cm} \Large \textbf{An Empathetic AI Coach for Self-Attachment Therapy\vspace{0.4cm}}\\
\thanks{Neophytos Polydorou was supported by the UKRI CDT in AI for Healthcare http://ai4health.io (Grant No. P/S023283/1).}}

\author{\IEEEauthorblockN{Lisa Alazraki\IEEEauthorrefmark{1},
Ali Ghachem\IEEEauthorrefmark{1},
Neophytos Polydorou\IEEEauthorrefmark{1}, 
Foaad Khosmood\IEEEauthorrefmark{2} and
Abbas Edalat\IEEEauthorrefmark{1}}
\IEEEauthorblockA{\IEEEauthorrefmark{1}\textit {Department of Computing}\\
\textit {Imperial College London, United Kingdom}\\ \textit{Email: \{lisa.alazraki20, ali.ghachem17, neophytos.polydorou19, a.edalat\}@imperial.ac.uk}}
\IEEEauthorblockA{\IEEEauthorrefmark{2}\textit{Computer Science and Software Engineering}\\
\textit{California Polytechnic State University,
San Luis Obispo, USA}\\
\textit{Email: foaad@calpoly.edu}}}


\maketitle

\begin{abstract}
In this work, we present a new dataset and a computational strategy for a digital coach that aims to guide users in practicing the protocols of self-attachment therapy. Our framework augments a rule-based conversational agent with a deep-learning classifier for identifying the underlying emotion in a user's text response, as well as a deep-learning assisted retrieval method for producing novel, fluent and empathetic utterances. We also craft a set of human-like personas that users can choose to interact with. Our goal is to achieve a high level of engagement during virtual therapy sessions. We evaluate the effectiveness of our framework in a non-clinical trial with N=16 participants, all of whom have had at least four interactions with the agent over the course of five days. We find that our platform is consistently rated higher for empathy, user engagement and usefulness than the simple rule-based framework. Finally, we provide guidelines to further improve the design and performance of the application, in accordance with the feedback received.
\end{abstract}


\fontsize{10}{11}\selectfont

\section{Introduction}

It is estimated that almost a billion people worldwide -- approximately 13 percent of the global population -- suffer from at least one mental disorder~\cite{collaborators_2018}. This number has increased by a third since 1990, and it is expected to continue to grow at an even steeper rate in the near future, due to the direct and indirect effects of the COVID-19 pandemic~\cite{Abbott2021}. Despite the demonstrated need for pervasive, affordable mental healthcare, the considerable personal financial cost that is often associated with traditional psychotherapy prevents patients from low-income backgrounds from accessing therapy~\cite{krupnick12}. Moreover, patients in low and middle-income countries and rural areas encounter a further barrier to accessing specialised care, due to low local ratios of mental health professionals per capita~\cite{Fu2020, weightman2020}. Confronted with these issues, researchers have examined digital technology as a means to deliver mental health services to the wider population~\cite{Renn2019}. As a result, a wide range of technological tools aimed at mental health support has been investigated and deployed within academia and industry~\cite{fairburn2017, blumenfield2020}, many of which take the form of conversational agents administering various forms of psychotherapy~\cite{gaffney2019}.

It should be noted, however, that using conversational agents in a sensitive area such as mental healthcare poses significant challenges. Current deep-learning approaches to text and speech generation lack the necessary oversight to prevent a system from producing output that is insensitive~\cite{Miner2016SmartphoneBasedCA} and even offensive~\cite{henderson18}, and thus potentially damaging to a patient’s well-being. A recent literature review study has observed that the large majority of mental-health-oriented chatbots currently in existence do not use machine learning at all, favouring more stable and predictable techniques such as rule-based modelling~\cite{alrazaq2019}. On the other hand, purely rule-based bots have a limited, keyword or pattern-based understanding of user input and their dialogue can be perceived as monotonous and predictable~\cite{Hussain19}, resulting in a failure to fully engage users.

In this paper, we present a computational framework that augments a rule-based agent for the delivery of self-attachment technique (SAT), a recently developed psychotherapeutic intervention~\cite{Edalat2017}. Our approach is aimed at maintaining the safety of rule-based strategies while also ensuring that the conversational agent generates responses that are empathetic, diverse and fluent, as well as appropriate to the user's emotional state. To this end, we create a new dataset -- E{\footnotesize MPATHETIC}P{\footnotesize ERSONAS} -- of 1,181 verbal expressions of emotion and 2,143 empathetic rewritings of base utterances, both crowd-sourced. We adopt a tree-structured conversation flowchart and devise a strategy for generating, at each node in the chart, novel yet safe utterances, trying to minimise any unpredictability in their overall meaning. To do so, we extract short, self-contained sentences from the set of utterance rewritings in the E{\footnotesize MPATHETIC}P{\footnotesize ERSONAS} dataset, by splitting each of them at major punctuation marks. We then join the extracted sentences in all possible sequential combinations and obtain a large corpus of new utterances. From this corpus, the agent retrieves -- through a multi-objective function that simultaneously maximises empathy, fluency and novelty -- the most appropriate utterance to present to the user. To compute the empathy score of an utterance, we use a T5 model~\cite{roberts2020} that is fine-tuned on a labelled subset of our dataset ($\sim$80\% accuracy, $\sim$81\% macro F1); for the fluency score we subtract a penalty for each repeated word within an utterance from the inverse of its perplexity generated by a GPT-2 language model~\cite{Radford2019LanguageMA}; finally, to obtain the novelty score, we compute a weighted overlap distance over all possible $n$-grams between an utterance and each of the agent's previous utterances. In addition, we adopt a RoBERTa model \cite{liu2020roberta} for the task of emotion recognition ($\sim$95\% accuracy, $\sim$95\% macro F1) that is trained on an existing affective dataset~\cite{saravia-etal-2018-carer} and further fine-tuned on the expressions of emotion in our corpus. This allows the bot to identify a user's emotional state from their text responses and answer accordingly. Lastly, we craft human-like characters for our conversational agent which users can choose from and interact with. Our dataset and code are publicly available \cite{SATbot}.

We evaluate the application through a human trial with N=16 subjects from the non-clinical population, as well as two medical professionals specialised in mental health. We show that our approach is scored highly for perceived usefulness, ability to communicate empathetically and user engagement, and that it performs significantly better than the simple rule-based version of the SAT chatbot~\cite{Ghachem2021} in all three areas. Our agent's ability to recognise human feelings is also assessed positively, with 63\% of trial participants agreeing that the bot was successful in guessing their emotions. In light of the feedback received during the trial, we conclude with a reflection on the strengths of our work as well as the weaknesses, drawing a list of changes and improvements which we believe may benefit the chatbot and its users.

\section{Background}

\vspace{-3.5pt}

\subsection{Existing approaches to chatbot-assisted mental support}

Many of the mental health support chatbots currently in existence approach dialogue generation using a tree-structured flowchart, whose transitions between prearranged states are determined by user input~\cite{kraus21, Denecke20, Ly2017, morbini-etal-2012-mixed, Bauer2020MeTooMaastrichtBA, Fitzpatrick2017, ghandeharioun19, Ali_2020}. The input can take the form of open text~\cite{Bauer2020MeTooMaastrichtBA, Fitzpatrick2017}, multiple choice~\cite{kraus21} or a combination of the two \cite{Denecke20, Ly2017, morbini-etal-2012-mixed}. Within this framework, the conversation can be modelled as a slot-filling problem, where the user's input is integrated into pre-existing templates to create a chatbot utterance~\cite{Bauer2020MeTooMaastrichtBA, Ali_2020, Fitzpatrick2017}. Alternatively, it can be informed by completely fixed, predetermined utterances, often written by mental health professionals with formal psychology training~\cite{Denecke20, Ly2017}. Using fixed templates and utterances enables researchers to maintain control over the dialogue, ensuring that the bot will not deliver insensitive or problematic responses which could potentially have a negative effect on the patient's mental health. However, this can also render the experience less engaging due to the conversation appearing rigid and repetitive, especially if a user interacts with the chatbot on a regular basis~\cite{boringbots}. To introduce a degree of variety in the conversation, Ghandeharioun et al.~\cite{ghandeharioun19} propose a retrieval method that randomly selects each bot utterance from a set of variations; however, the set only comprises six options, and thus it is unlikely to be able to prevent the dialogue from becoming repetitive over time.

\subsection{Digital psychotherapy and self-attachment technique}

Self-attachment technique (SAT) is a recently developed psychotherapy framework consisting of 20 self-administered protocols~\cite{edalatSAT} aimed at establishing and reinforcing neural patterns associated with secure attachment~\cite{Edalat2017}. It stems from findings in developmental psychology that link insecure attachment of children with their primary caregivers with affective disorders in adulthood~\cite{Mikulincer2012}. In SAT, the patient simultaneously enacts the roles of the adult -- corresponding to the logical self -- and the child -- representing the emotional self -- gradually building a bond between the two. The adult self re-parents the childhood self by emulating the optimal interactions of a real parent with their child. This allows the childhood self to become securely attached to the adult self, enhancing positive emotions and equipping the patient with the cognitive tools to tackle challenging situations and negative feelings. SAT can be used to alleviate mental illness and it can also increase social and emotional learning in the normal population. 

SAT is suitable to be dispensed in a digital, automated manner due to its self-administered nature. A virtual reality (VR) platform for SAT has been developed in both a high and a low-end version~\cite{GGNE20}. The high-end VR platform, based on Facebook's Oculus, has also been equipped with an audio-based emotion recognition system and a dialogue manager~\cite{PE21}. In addition, a recent study investigating the applicability of a chatbot for the delivery of SAT received some encouraging results, with 40\% of participants rating the platform as useful~\cite{Ghachem2021}. On the other hand, the entirely rule-based bot was deemed to be empathetic by only 20\% of respondents, while 30\% agreed that conversing with it was an engaging experience. Here, we extend the previous work done on the SAT chatbot by leveraging deep learning methods for emotion recognition and utterance retrieval. Our goal is to increase users' perception of empathy and overall engagement.

\subsection{Empathy in digital psychotherapy}

According to psychotherapy research, the most important factor to ensure the establishment of a beneficial relationship between a therapist and their patient is the ability of the former to engage in an empathetic manner with the latter~\cite{Elliott2018}. We thus consider empathy to be an indispensable feature for a mental health support chatbot. We adopt the definition of empathy given by Barrett-Lennard~\cite{Barrett1981}, who identifies three main phases of an empathetic dialogue between two individuals: a first phase where the listener sympathises and resonates with what is being expressed by the speaker, a second phase in which the listener compassionately responds to the speaker, and a third phase where the speaker assimilates the listener's response. Here we mainly focus on Barrett-Lennard's second phase -- the \textit{expressive} phase of empathetic exchange -- in an attempt to create a digital psychotherapist able to display compassion toward the user.

\subsection{Chatbot personification and user engagement}

Past research shows that users' experience of interacting with a chatbot improves significantly when this is equipped with a coherent identity \cite{Kocaballi2019}. Moreover, psychology studies have highlighted that individuals tend to prefer psychotherapists of a certain age or sex according to several factors. For example, women generally report higher levels of comfort when self-disclosing to female practitioners compared to male ones~\cite{Landes2013}, and patients tend to choose younger or older specialists depending on the specific issue that they are facing (older therapists are preferred for universal problems such as mourning, while younger ones are favoured when dealing with issues that more typically affect young people, such as heartbreak or cyberbullying)~\cite{Kessler2020}. In an attempt to increase users' engagement with the conversational agent, we create for it a set of personas characterising different sexes and age ranges. Section~\ref{ee} includes details of how these personas are created.

\vspace{-2pt}

\subsection{Privacy and ethics}

User input saved during interactions to provide the service is permanently deleted at the end of each session. The application does not collect or store geolocation data, IP or MAC addresses or any other metadata from users' devices.

It should be noted that our chatbot is designed for individuals who are already familiar with SAT and has not yet been tested on the clinical population. In its present form, the bot could produce responses that may be inappropriate within contexts involving self-harm. A careful and considered approach should be taken when dealing with users that may be experiencing mental distress, and future research should meticulously assess any risks associated with using the platform in a clinical setting as well as the appropriate solutions.

\subsubsection*{Ethical approval}

In this work, the collection of the E{\footnotesize MPATHETIC}P{\footnotesize ERSONAS} dataset and the non-clinical trial for the evaluation of the SAT chatbot have been approved by the Research Ethics Committee of Imperial College London.

\section{Dataset and data collection}

\subsection{Survey preparation }
 
We crowd-sourced the E{\footnotesize MPATHETIC}P{\footnotesize ERSONAS} dataset by distributing four surveys. Each survey contained two tasks: one asking respondents to provide multiple textual expressions of emotion (answering the question `How are you feeling?') for different emotional contexts, and one requiring them to rewrite a set of base utterances to render them empathetic, keeping in mind that these utterances are to be directed to an interlocutor who is experiencing a specified emotion. In addition, we asked respondents to provide information about their sex and age.

\begin{figure}
\centering
\includegraphics[width=\linewidth]{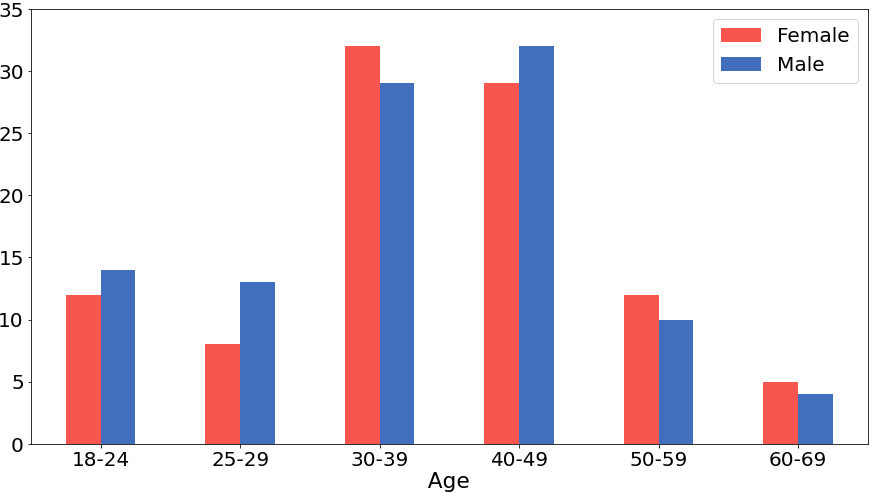}
\caption{\hbadness=10000 Age distribution for both sexes across samples in the E{\scriptsize MPATHETIC}P{\scriptsize ERSONAS} dataset, showing that most samples belong to the middle age groups 30-39 and 40-49.}
\label{fig1}
\vspace{-4pt}
\end{figure}

\vspace{-2pt}

\subsection{Recruitment of survey respondents}

Survey respondents were recruited via the crowd-working websites Amazon Mechanical Turk \footnote{https://www.mturk.com} and Prolific \footnote{https://prolific.co}. All recruited respondents were educated at college level or above and their first language was English.

\subsection{Criteria for the acceptance of responses}

Responses were rejected if they amounted to less than 50 percent of the survey, if they contained poorly written syntax or unrelated text, or if the base utterances that were meant to be rewritten had been copy-pasted without changes. In all the other cases, the responses were accepted. Where minor grammar, syntax or semantic mistakes were present, these were rectified before insertion into the dataset. We modified the punctuation in some of the empathetic utterance rewritings by replacing commas with full stops whenever these were positioned at the end of a complete sentence. In total, 200 responses were accepted -- 50 for each of the four surveys.

\subsection{Data analysis}

The E{\footnotesize MPATHETIC}P{\footnotesize ERSONAS} dataset comprises 200 rows, each corresponding to a survey response. Each row contains the sex and age range of the respondent, as well as the expressions of emotions and empathetic rewritings that they provided. There are two sexes (male, female) and six age groups within the corpus. While the distribution of data samples across the two sexes is balanced (98 females and 102 males), the majority of the samples originate from the 30-39 and 40-49 age groups for both sexes, as shown in Fig.~\ref{fig1}. 

The dataset contains 1,181 textual expressions of emotion distributed across four emotional contexts: 299 are expressions of sadness, 297 communicate anger, 285 relate to anxiety/fear and 300 convey happiness/content. It also includes 2,143 empathetic rewritings of 45 base utterances. Each subset of 50 rows collects the responses to one of the four surveys and contains different emotional contexts as well as rewritings of different base utterances. Accounting for some missing data, the corpus comprises between 42 and 50 rewritings of each base utterance. All empty cells are filled with NaN values.

\subsection{Creating personas from the data}\label{ee}

We used the information collected about the sex and age group of the survey respondents to divide the data into four subsets, each of which informs a different chatbot persona. Therefore, the empathetic rewritings provided by female crowd-workers aged 18 to 39 condition the dialogue of a younger female persona named Olivia, while those provided by male respondents in the same age range inform the conversation of a younger male persona named Arman. Similarly, we created an older female persona named Gabrielle, whose dialogue is based on the rewritings provided by female crowd-workers aged 40 to 69, and an older male persona named Robert, whose interactions are crafted from the survey responses given by male crowd-workers aged between 40 and 69. We also created a further identity for our chatbot named Kai, whose dialogue is informed by the whole dataset and is not associated to any sex or age group.

\subsection{Empathy annotation}\label{empathyann}

The utterance rewritings in our corpus may convey different degrees of empathy. This is due to the individual personality of each survey respondent and their interpretation of the task, as well as the fact that we did not reject responses based on their perceived degree of empathy. In order to build an effective empathy classifier, necessary to ensure that our system produces the most appropriate responses, we created a separate dataset by randomly extracting 1,100 utterance rewritings from the corpus and annotating them for empathy, using discrete numerical labels from 0 to 2 (where 0 corresponds to a non-empathetic utterance and 2 to a strongly empathetic one). We used this scale as it correlates with previous work in automated empathy recognition~\cite{Sharma2020}. To avoid excessively biasing the model toward our own judgement, we enlisted two volunteers to re-annotate the 1,100 rewritings for empathy, using the same scale. Both annotators have worked in healthcare and are experienced in communicating empathetically with patients. For each rewriting, we computed the overall empathy score by choosing the majority label out of the three individual ones. If all three labels were different, we assigned a score of 1.

It should be noted that this labelling method may still invite bias, as all three annotators belong to similar age groups (30-39 and 40-49). In future implementations, it is recommended that the rewritings are re-scored via crowd-sourcing.

\section{Implementation}

\subsection{Emotion recognition}\label{emorec}

To customise the dialogue to the relevant emotional context, the chatbot asks the user to describe how they feel at the beginning of each conversation. Consistently with the data collected in the E{\footnotesize MPATHETIC}P{\footnotesize ERSONAS} dataset, we aim to discern between four contexts: sadness, anger, anxiety/fear and happiness/content. To achieve effective emotion recognition given a user's text response, we fine-tuned a pretrained RoBERTa language model for this task, first on Saravia et al.'s affective dataset~\cite{saravia-etal-2018-carer} and then on the expressions of emotion in our corpus. The model achieves 94.96\% accuracy and 95.10\% macro-averaged F1 score on the test set split from our corpus (in contrast, the keyword-based emotion classifier implemented in the previous version of the chatbot~\cite{Ghachem2021} obtains 63.03\% accuracy and 62.48\% macro F1 on the same test set). Table~\ref{tab1} displays the hyperparameters used in both fine-tunings.

It should be noted that the expressions of emotion in the E{\footnotesize MPATHETIC}P{\footnotesize ERSONAS} dataset have been provided by individuals instructed to answer the question `How are you feeling?' \textit{as if} they were experiencing a particular emotion. The fact that our model's second and final fine-tuning was not performed on genuine emotional expressions -- but rather on their imitation -- is potentially a source of bias that may decrease its performance when applied to real-world situations.

\renewcommand{\arraystretch}{1.5}
\hbadness=10000
\begin{table}
\caption{Hyperparameters used to fine-tune a RoBERTa language model for the task of emotion recognition.}
\begin{center}
\begin{tabular}{|p{1.65cm}|p{1.55cm}|p{1.15cm}|p{0.7cm}|p{1.7cm}|}
\hline 
\textbf{Train-val-test proportions} & \textbf{Learning rate} & \textbf{Adam epsilon} & \textbf{Batch size} & \textbf{ Epoch with best accuracy} \\
\hline
$80$-$10$-$10$ & $1.35 \times 10^{-4}$ & $1 \times 10^{-8}$ & $16$ & $10$ \\
\hline
\end{tabular}
\label{tab1}
\end{center}
\end{table}

\begin{figure}
\centering
\includegraphics[width=\linewidth]{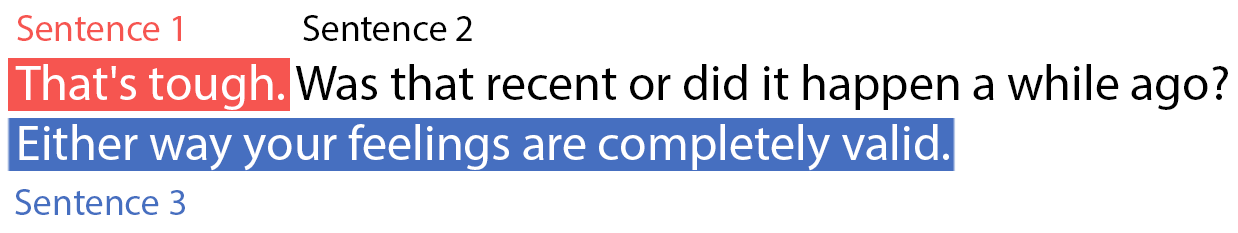}
\caption{A three-sentence utterance rewriting in the E{\scriptsize MPATHETIC}P{\scriptsize ERSONAS} dataset. Sentence 2 conveys the main question, while Sentences 1 and 3 reinforce the empathy of the message by expressing sympathy and compassion.}
\label{sentence}
\vspace{-4pt}
\end{figure}

\subsection{Corpus augmentation}\label{bb}

\begin{figure}
\centering
\includegraphics[width=\linewidth]{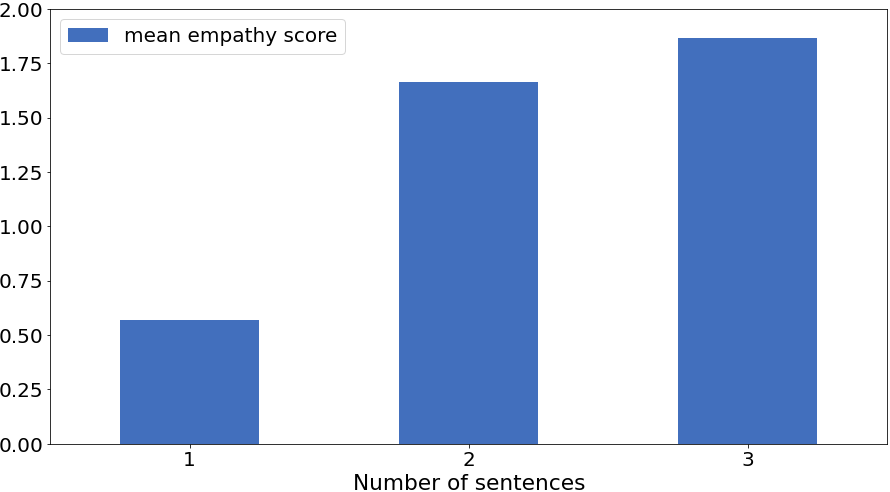}
\caption{\hbadness=10000 Bar chart showing the mean empathy score of rewritten utterances in the E{\scriptsize MPATHETIC}P{\scriptsize ERSONAS} dataset by number of sentences that they contain (for details of the empathy annotation process see Section \ref{empathyann}).}
\label{sentenceanalysis}
\end{figure}

The utterance rewritings in the E{\footnotesize MPATHETIC}P{\footnotesize ERSONAS} dataset consist of either one, two or three distinct sentences. Fig.~\ref{sentence} illustrates an example of a three-sentence rewriting in the corpus. We therefore extracted individual sentences from the dataset by splitting each rewritten utterance at major punctuation marks (full stops, questions marks and exclamation points), and recombined these sentences together in different ways to  form new utterances. This approach has the following advantages: (a) it allows the augmentation of our text data, otherwise bound to the limited size of the dataset; (b) it ensures that the newly-generated utterances remain safe and reliable, since each sentence is self-contained in its meaning, has been reviewed at the dataset collection stage and is known not to be insensitive or harmful; (c) it has the potential to increase the level of empathy of those rewritten utterances which may not be highly empathetic in their original form. As shown in Fig.~\ref{sentenceanalysis}, further analysis of our data shows that utterances composed of two or more sentences are perceived on average as more empathetic by human annotators compared to single-sentence ones. This may be due to the fact that, when an utterance is composed of several sentences, one of them conveys the main message while the others are often expression of politeness, sympathy or compassion.

When extracting sentences, we aimed to save a record of their relative position within the original utterance in order to maintain this position when combining them together to form new utterances, thus increasing the likelihood of a meaningful result. We defined three lists -- $first\_pos\_list$, $second\_pos\_list$ and $third\_pos\_list$ -- corresponding to the three possible positions within an utterance (since the utterances in our corpus contain at most three sentences), and assigned each extracted sentence to one of these lists. Of course, the assignment is straightforward when utterances are composed of three sentences, whereas for shorter utterances we employed a strategy to achieve a sensible assignment \cite{Alazraki2021}, populating $second\_pos\_list$ with the sentences most likely to convey the main message of an utterance. Having populated the three lists, we eliminated from them any duplicate sentences and added an empty string to $first\_pos\_list$ and $third\_pos\_list$ (but not to $second\_pos\_list$, to prevent the possibility of creating empty utterances by selecting the empty string from all three position lists). We then formed new utterances containing one, two or three sentences by successively choosing one item from each of $first\_pos\_list$, $second\_pos\_list$ and $third\_pos\_list$, in this order, until all possibilities had been exhausted. The resulting corpus thus contains $|\,first\_pos\_list\,|$ $\times \: |\,second\_pos\_list\,| \: \times \: |\,third\_pos\_list\,|$ utterances (where the notation $|\,list\,|$ indicates the length of $list$). This process was repeated for each column in the  E{\footnotesize MPATHETIC}P{\footnotesize ERSONAS} dataset (i.e. we only combined together sentences originating from rewritings of the same base utterance).

Through the process of sentence extraction and recombination we obtained corpora of utterances significantly larger than the original, as illustrated in Table~\ref{size_comparison}. Visual inspection of these corpora reveals that the quality of the newly-generated utterances is, on average, satisfactory. However, not all utterances are equally suitable to be used by the chatbot. Some of them may be less fluent than others, due to repetitions or semantic conflicts arising from combining parts of different rewritings, and some may still lack enough empathy. Moreover, many utterances have sentences in common, increasing the risk that the bot's dialogue may sound repetitive. To overcome these issues, we devised a retrieval method that yields the best possible utterance at each stage of the conversation.

\subsection{Retrieval method}

Our retrieval method consists of a multi-objective optimisation function combining an empathy score, a fluency score and a novelty score, which are simultaneously maximised when selecting an utterance.

\subsubsection*{Empathy function} \label{empathyClassifier}

To compute the empathy score of an utterance, we fine-tuned a T5 language model on the portion of the E{\footnotesize MPATHETIC}P{\footnotesize ERSONAS} dataset that had been annotated for empathy (see Section~\ref{empathyann}). We obtained a classification accuracy of 80.18\% and a macro-averaged F1 score of 80.66\% on the test set. Table~\ref{tab3} illustrates the hyperparameters used in the fine-tuning process. The values returned by this model, which corresponds to our empathy scoring function $E$, are normalised to be between 0 and 1 by dividing each output by 2 (which is the maximum empathy score possible).

\begin{table}
\caption{Comparison of the total number of utterances in each dataset split before and after the augmentation process.}
\begin{center}
\begin{tabular}{|p{3.05cm}|p{2.35cm}|p{2.2cm}|}
\hline 
\textbf{Dataset split and} &\multicolumn{2}{|c|}{\textbf{Total number of utterances}} \\
\cline{2-3} 

\textbf{associated persona} & \textbf{\textit{Before augmentation}} & \textbf{\textit{After augmentation}} \\ \hline

{Males 40-69 (Robert)}  & {480} & {\footnotesize 3,980} \\ \hline

{Females 40-69 (Gabrielle)} & {495} & {4,123} \\ \hline

{Males 18-39 (Arman)} & {614} & {4,747} \\ \hline

{Females 18-39 (Olivia)} & {554} & {5,172} \\ \hline

{Entire dataset (Kai)} & {2,143} & {94,993} \\ \hline

\end{tabular}
\end{center}
\label{size_comparison}
\end{table}

\begin{table}
\caption{Hyperparameters used to fine-tune a T5 language model for the task of empathy classification.}
\begin{center}
\begin{tabular}{|p{1.65cm}|p{1.55cm}|p{1.15cm}|p{0.7cm}|p{1.7cm}|}
\hline
\textbf{Train-val-test proportions} & \textbf{Learning rate} & \textbf{Adam {} epsilon} & \textbf{Batch size} & \textbf{Epoch with best accuracy} \\
\hline
$80$-$10$-$10$ & $1 \times 10^{-4}$ & $1 \times 10^{-8}$ & $8$ & $16$ \\
\hline
\end{tabular}
\label{tab3}
\end{center}
\vspace{-5pt}
\end{table}

\subsubsection*{Fluency function}

To evaluate the fluency of an utterance, we compute the inverse of its perplexity (\textit{PPL}) score returned by a GPT-2 language model. Since combining together portions of different utterances may create unwanted repetitions, we subtract from this value a penalty of 10$^{-2}$ for each repeated (lemmatised) word, excluding stop words. Therefore, the fluency $F$ of an utterance $u$ is given by
\begin{equation}
F(u) =  \mfrac{1}{PPL (u)} - RP(u), \label{eq1}
\end{equation}
where $RP(u)$ is the total penalty for all the repeated words within that utterance. To normalise \eqref{eq1} so that it returns values through the whole range between 0 and 1, we divide it by the maximum possible fluency score as calculated on the augmented corpora (i.e. 0.16). If the output is negative, which may happen when the total penalty is greater than the inverse of the perplexity, we return zero.

\subsubsection*{Novelty function}

The chatbot is able to save and retrieve up to 50 of its previous utterances, and it compares each new utterance to those in this set to evaluate its novelty. To this end, we implement a function that computes the weighted overlap distance~\cite{Vijaymeena2016} over all possible $n$-grams between two text sequences, starting from unigrams up to $N$-grams where $N$ is equal to the length in words of the shorter sequence. The greater the number $n$ the more we decrease the distance between $n$-grams -- which is a number between 0 and 1 -- by raising it to the power $n$, since utterances are more similar when they share longer sequences of words. After adding together the distances over all possible $n$-grams, we divide the result by $N$ so that it remains between 0 and 1. The distance $d$ between two utterances $u_1$ and $u_2$ is thus given by

\begin{equation}
d(u_1, u_2) = \frac{ \sum_{n=1}^{N} \left( 1 - \frac{|n\textrm{-grams}(u_1) \: \cap \: n\textrm{-grams}(u_2)|}{\textrm{min}(|n\textrm{-grams}(u_1)|, |n\textrm{-grams}(u_2)|)} \right)^n}{N},\label{eq2}
\end{equation}
where $n$-grams$(u)$ represents the set of $n$-grams in the utterance $u$ and the notation $|X|$ indicates the size of set $X$. Equation~\eqref{eq2} is computed between a new utterance and each of the saved previous utterances, adding up the results to obtain the novelty (or diversity) score $D$ of the new utterance. We divide $D$ by the number of previous utterances to obtain a normalised value between 0 and 1.

\subsubsection*{Multi-objective optimisation function}\label{multiobj}

Let $E_{norm}(u)$, $F_{norm}(u)$ and $D_{norm}(u)$ be the normalised functions measuring, respectively, the empathy, fluency and diversity of an utterance $u$, each returning a value between 0 and 1. Then, the overall function $R$ that we wish to maximise when retrieving a new utterance is given by 

\begin{equation}
R(u) = w_e E_{norm}(u) + w_f F_{norm}(u) + w_d D_{norm}(u). \label{eq3}
\end{equation}
We fix the weights in \eqref{eq3} to $w_e = 1$, $w_f = 0.75$ and $w_d = 2$. These values have been obtained experimentally to give reasonable results. It should be noted that calculating $R(u)$ is computationally expensive: the complexity of transformer-based models such as T5 and GPT-2 -- which we use to compute $E_{norm}(u)$ and $F_{norm}(u)$ respectively -- is quadratic in the length (in words) of the utterance $u$~\cite{NIPS2017_3f5ee243}. Moreover, the function $D_{norm}(u)$ performs for each new utterance $p \times N \times (N+1)/2$ comparisons, where $p$ is the number of saved previous utterances and $N$ is the length, in words, of the shorter of the two utterances being compared. As a trade-off between response time and size of the utterance retrieval pool, we apply \eqref{eq3} on a random subset of 15 utterances drawn from the corpus. In future implementations, it may be worth precomputing the empathy and fluency scores of each utterance and appending these values to the augmented corpora, so that only the novelty score, which depends on the bot's previous utterances, will need to be calculated at runtime.

\subsection{Conversation flow}

\begin{figure}
\centering
\includegraphics[width=0.9\linewidth]{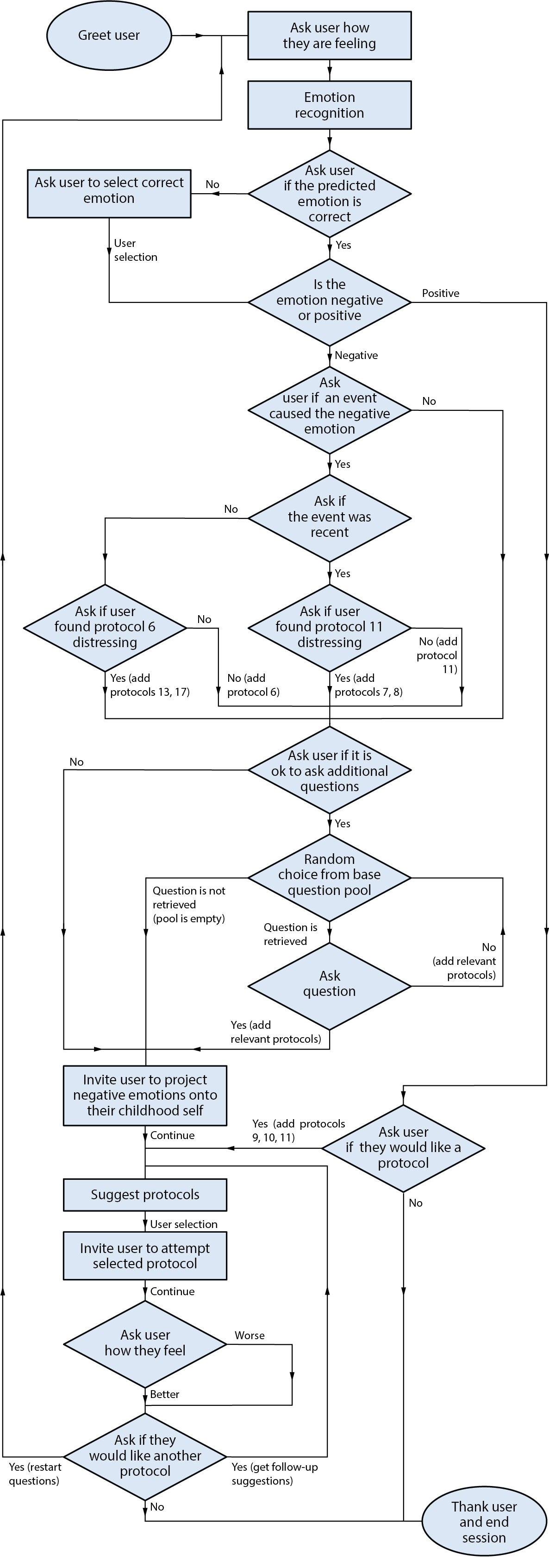}
\caption{Conversation flow of the SAT chatbot.}
\label{flowchart}
\end{figure}

After a user has logged into the application, the bot asks them to choose a persona between Kai, Robert, Gabrielle, Arman and Olivia. The user's selection informs which portion of the (augmented) data is loaded into the back-end. The conversation between the user and the chatbot allows a mix of open text and multiple-choice input and is informed by a flowchart, illustrated in Fig.~\ref{flowchart}. At each node in the flowchart, deep-learning methods are applied for emotion recognition or utterance retrieval. All five personas navigate the same flowchart when conversing with the user, but each of them has a specific set of utterances that they can retrieve from. Similarly, once a user's emotional state has been identified, it is saved as a variable and used to select relevant subsets of utterances from the dataset.

The main objective of the chatbot is to recommend the most appropriate SAT protocols. As users navigate the conversation, a list of protocol suggestions is drawn. The contents of this list, as well as the point in the dialogue where the bot discloses the suggestions, depend on the answers provided by the user.

\subsection{User interface}

\begin{figure}
\centering
\includegraphics[width=\linewidth]{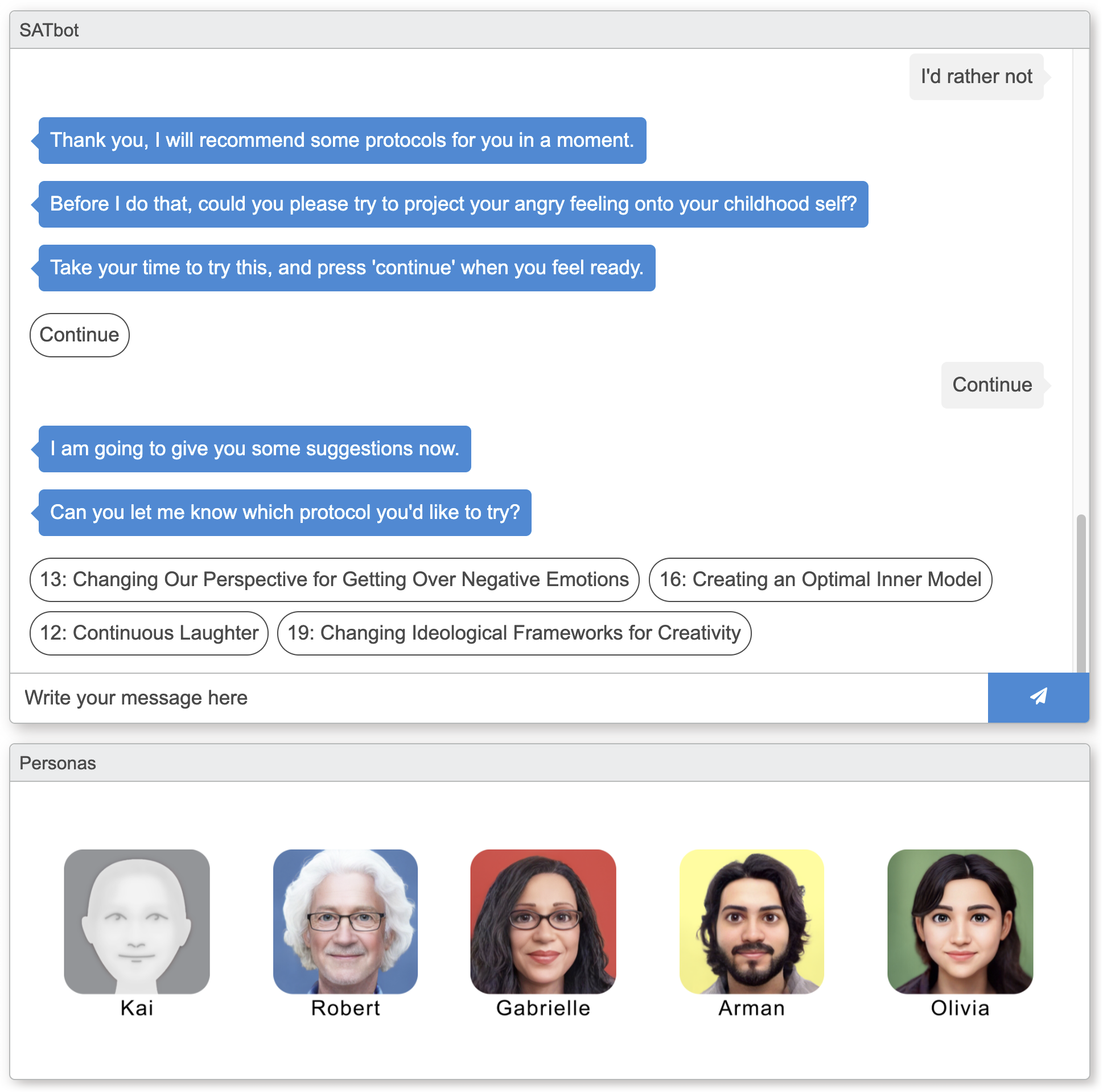}
\caption{Appearance of the SAT chatbot web application.}
\label{ui}
\end{figure}

The communication between the Python back-end and the JavaScript front-end is managed by the Flask API and the React-chatbot-kit library~\cite{reactkit}. We design avatars for the chatbot personas that users can choose from. Fig.~\ref{ui} shows the interface of the application set up for the evaluation trial.

\section{Non-clinical trial}

\subsection{Study design}

The SAT chatbot was formally evaluated through a human trial. The pool of participants comprised 23 volunteers from the non-clinical population aged between 22 and 70, all of whom were already familiar with SAT. Of these 23 individuals, 16 were male and 7 were female. Each volunteer agreed to have four interactions with the chatbot over the course of five days -- two with Kai and the rest with any two of the other personas. The chatbot was also evaluated separately by two clinicians specialised in mental health, who completed the same number of interactions as the other participants.

The chatbot platform was deployed as a web application and all the interactions occurred online. Participants were sent instructions, a link to access the platform and individual login credentials via e-mail, and they were able to give feedback by filling out an anonymous online questionnaire. The questionnaire contained multiple-choice questions asking to evaluate: (a) the chatbot's ability to display empathy; (b) the level of engagement of each user; (c) the usefulness of the platform; (d) the ability of the chatbot to identify emotions. When volunteers evaluated the bot for empathy and engagement, they scored these attributes separately for Kai and the other personas. By collecting this information, we aimed to assess whether a human-like character -- such as Robert, Gabrielle, Arman and Olivia -- can improve user experience. On the other hand, we gauged whether having a much larger pool of utterances to choose from (and thus potentially more diversity in the responses), as is the case for Kai, can provide a significant advantage. The questionnaire also asked volunteers to state which personas they had interacted with and there were additional open-ended questions to collect comments and suggestions.

\subsection{Evaluation}\label{eval}

Of 23 study volunteers, 16 returned a complete questionnaire. The evaluation in this section is thus carried out on a sample size of 16. We compare our results with those obtained in a previous evaluation trial~\cite{Ghachem2021} of the earlier implementation of the SAT chatbot, which we define as our baseline. 

\subsubsection*{Evaluation by trial volunteers}

\begin{figure}
\centering
\includegraphics[width=\linewidth]{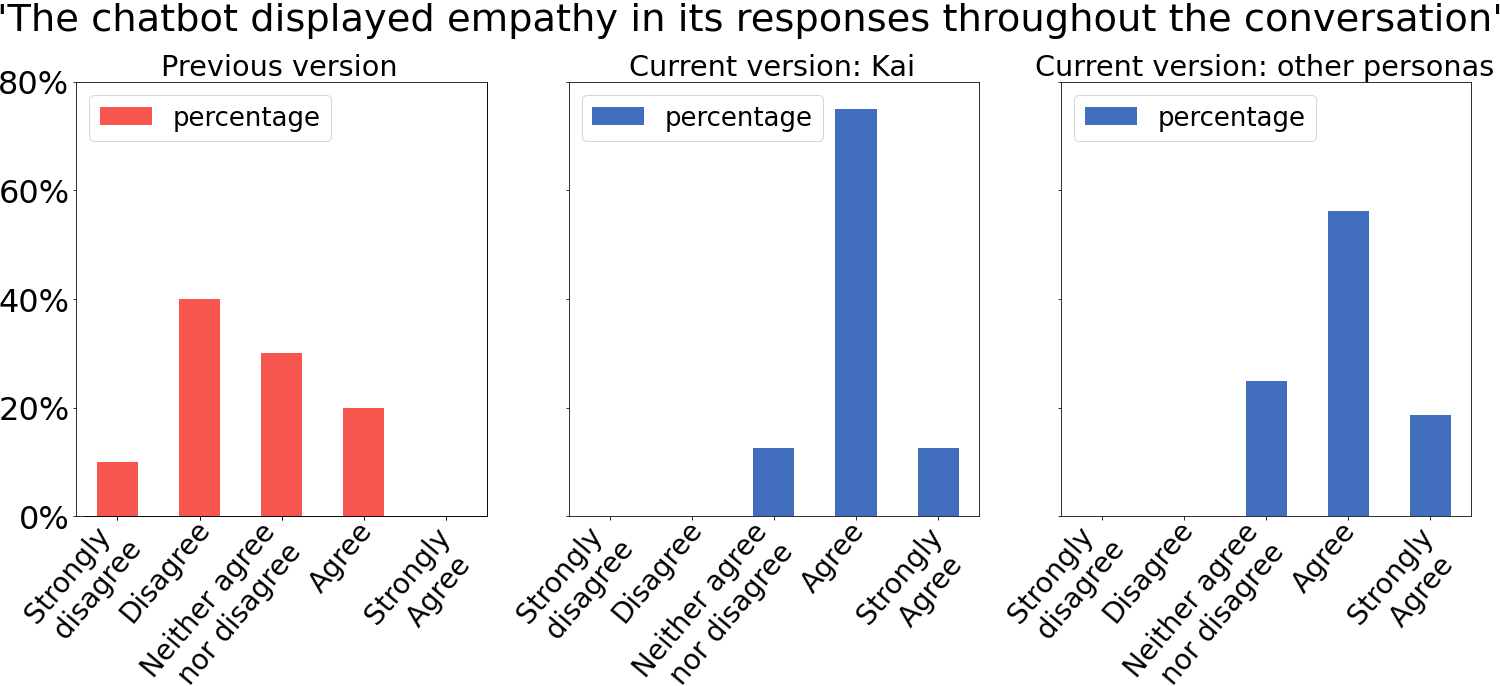}
\caption{Empathy evaluation of the previous version of the chatbot and the current one. Our results show significant improvement over the baseline in the perceived level of empathy for both Kai and the other personas.}
\label{empathyeval}
\end{figure}

Volunteers were asked to evaluate the chatbot's ability to convey empathy by expressing how much they agreed/disagreed with the statement `The chatbot displayed empathy in its responses throughout the conversation', both in the context of their interactions with Kai and in relation to the other personas. When interacting with Kai, 75\% agreed that the bot was empathetic, while the remaining quarter selected `Strongly agree' and `Neither agree nor disagree' in equal proportions, as illustrated in Fig.~\ref{empathyeval}. When the interactions were with any of the other personas, 56\% agreed with the statement, 19\% strongly agreed and a quarter neither agreed nor disagreed. In both cases we observe a significant improvement over the baseline: only 20\% of those participating in the previous trial agreed that the earlier implementation was empathetic, with 50\% expressing disagreement.

\begin{figure}
\centering
\includegraphics[width=\linewidth]{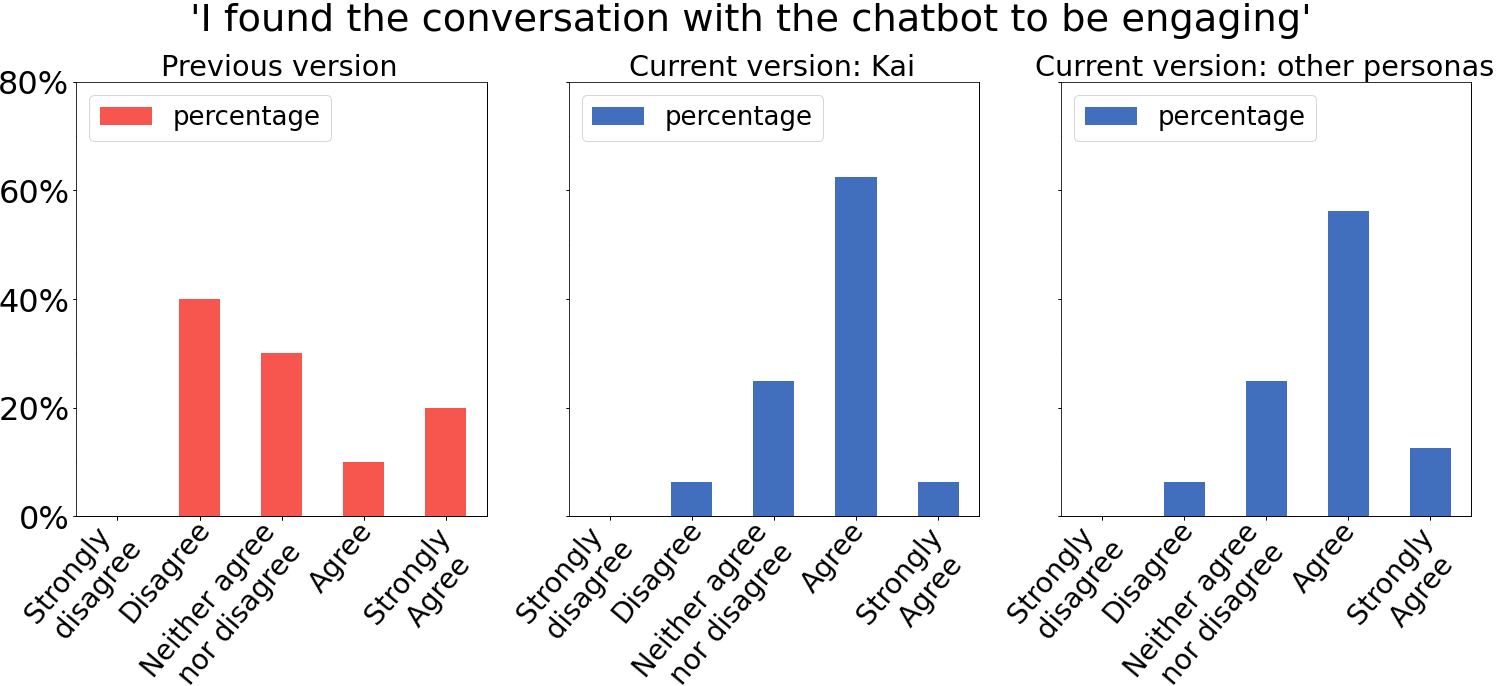}
\caption{Engagement level of users who interacted with the previous and current version of the chatbot. The level of user engagement improves in our implementation, whether the interactions are with Kai or the other personas.}
\label{engageval}
\end{figure}

When evaluating engagement, 6\% of participants disagreed with the statement `I found the conversation with the chatbot to be engaging', and a quarter neither agreed nor disagreed. This was the case for interactions with Kai as well as the other personas, as shown in Fig.~\ref{engageval}. In addition, 63\% agreed that Kai's conversations were engaging and a further 6\% strongly agreed, while 56\% agreed and 13\% strongly agreed that the other personas conversed in an engaging manner. In comparison, when evaluating the previous implementation, 40\% disagreed that the dialogue was engaging, 30\% neither agreed nor disagreed, and the remaining 30\% agreed or strongly agreed.

\begin{figure}
\centering
\includegraphics[width=0.75\linewidth]{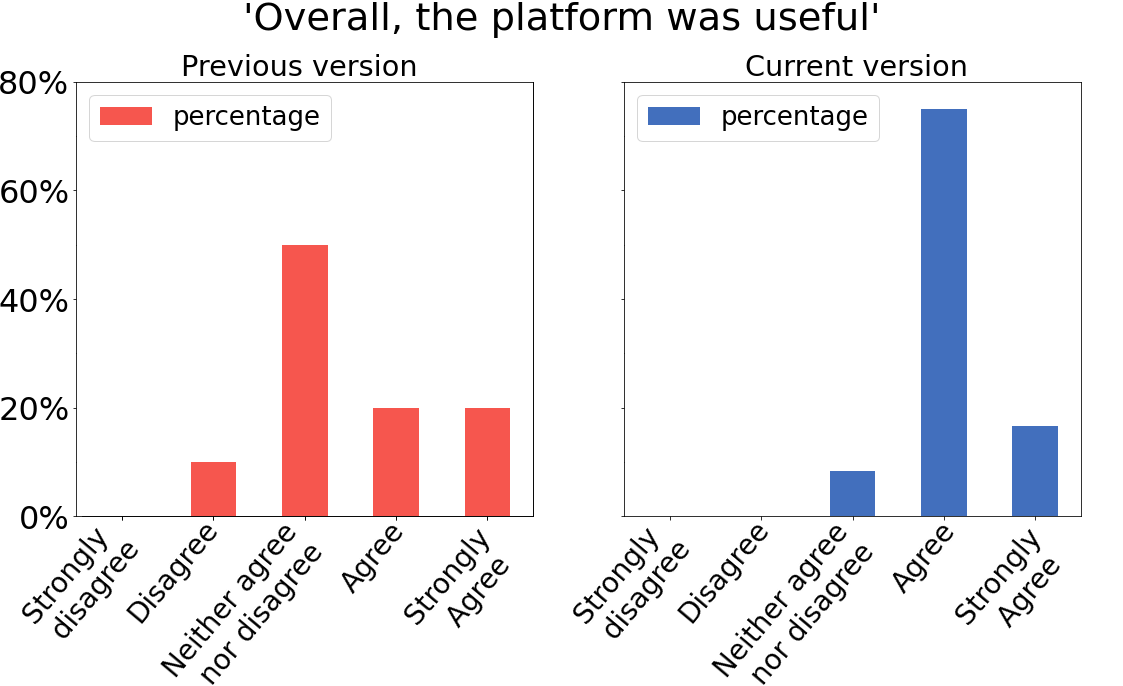}
\caption{Evaluation of usefulness of the earlier and current version of the chatbot, showing that the current version is more consistently rated as useful.}
\label{useval}
\end{figure}

Usefulness was evaluated by agreeing/disagreeing with the statement `Overall, the platform was useful'. Of our sample, 75\% agreed and a further 17\% strongly agreed with the above statement, with 8\% choosing `Neither agree nor disagree'. Fig.~\ref{useval} shows clear improvement over the baseline, which 10\% disagreed was useful, 50\% neither agreed nor disagreed, 20\% agreed and an equal proportion strongly agreed.

\begin{figure}
\centering
\includegraphics[width=\linewidth]{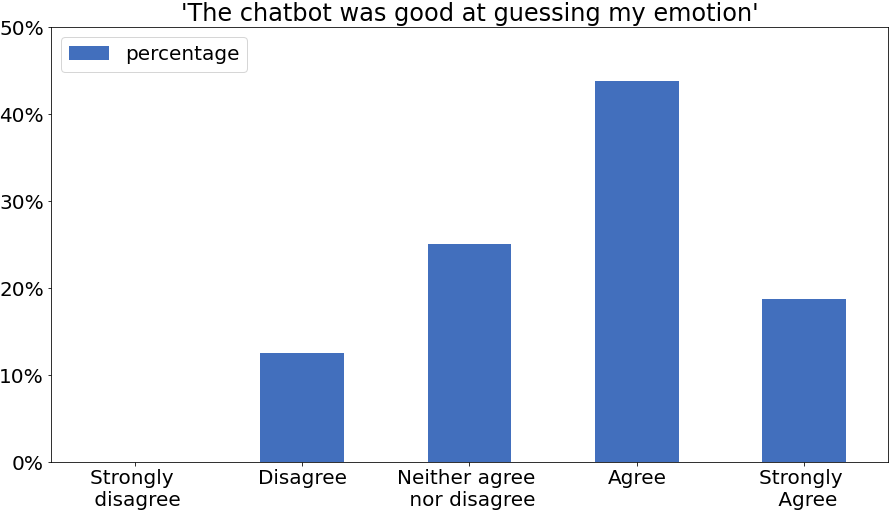}
\caption{Evaluation of the current SAT chatbot’s ability to recognise emotions.}
\label{emoeval}
\end{figure}

In addition, we found that 63\% of participants either agreed or strongly agreed with the statement `The chatbot was good at guessing my emotion', a quarter neither agreed nor disagreed and the remainder disagreed, as illustrated in Fig.~\ref{emoeval}. As no analogous data were collected during the previous trial, we cannot compare these results with the baseline. Instead, we refer the reader back to Section~\ref{emorec}, where our emotion recognition model is shown to achieve accuracy and macro-averaged F1 scores over 30\% greater than those obtained by the classifier used in the previous implementation (when tested on the same data).

Lastly, we investigated the volunteers' preferences when choosing personas. Without considering the mandatory interactions with Kai, we found that a quarter of the other interactions were with Olivia, approximately 15\% were with Gabrielle, and the remaining 60\% were equally split between Robert and Arman.

We should also note the comments and suggestions received. Some participants observed that the set of identifiable emotions was too limited, and this may have affected the bot's ability to successfully predict emotional states. Further feedback highlighted the fact that not only the range of emotions was narrow, but those emotions may have been quite extreme compared to what members of the non-clinical population would normally experience. For example, feeling `slightly worried' would be cast by the current version of the bot as being `anxious/scared', whereas the two states are arguably rather different. Several volunteers also noted that having only two choices (`I feel better' and `I feel worse') for giving feedback after completing a protocol is too restrictive, and more nuanced options would be required.

\subsubsection*{Evaluation by clinicians}

The clinicians' assessment of the earlier and current SAT chatbot is shown in Table~\ref{clintable}. It is worth noting that both specialists rated the platform identically regardless of whether their interactions were with Kai or the other personas. While their evaluation remained mostly unchanged, our implementation was viewed as significantly more empathetic than the previous one by one clinician, having turned their response to the statement `The chatbot displayed empathy in their responses throughout the conversation' from `Disagree' to `Agree'. The specialists commented positively on the chatbot's ability to interpret emotions, but observed that this was limited by the narrow range of emotional contexts available. They also noted that more positive reinforcement would be desirable (e.g. a congratulatory message when a user logs into the platform) as well as recognition and appropriate management of any user input involving self-harm.

\begin{table}
\caption{Clinician evaluation of our chatbot against the baseline.}
\begin{center}
\begin{tabular}{|p{2.78cm}|p{2.38cm}|p{2.38cm}|}
\hline 
\textbf{Statement} &\multicolumn{2}{|c|}{\textbf{Response}} \\
\cline{2-3} 

\textbf{} & \textbf{\textit{Previous version}} & \textbf{\textit{Current version}} \\ \hline

{The  chatbot  was good at guessing my emotion}  & {Clinician 1: N/A  \newline Clinician 2: N/A} & {\footnotesize Clinician 1: Agree  \newline Clinician 2: Agree} \\ \hline

\raggedright
{The  chatbot  displayed  empathy  in  its  responses throughout the conversation}  & {Clinician 1: Disagree  \newline Clinician 2: Disagree} & {\footnotesize Clinician 1: Agree  \newline Clinician 2: Disagree} \\ \hline

{I found the conversation with the chatbot to be engaging} & {Clinician 1: Agree  \newline Clinician 2: Disagree} & {Clinician 1: Agree  \newline Clinician 2: Disagree} \\ \hline

\raggedright
{Overall, the platform was useful} & {Clinician 1: Agree  \newline Clinician 2: Agree} & {Clinician 1: Agree  \newline Clinician 2: Agree} \\ \hline

\end{tabular}
\end{center}
\label{clintable}
\end{table}

\section{Discussion}

\subsection{Results}

Our framework and study add to the existing body of knowledge in computational methods for mental health support. The human evaluation trial shows promising results with respect to the perceived empathy, user engagement, usefulness and ability to identify emotions of the chatbot. We find that trial participants report higher levels of engagement with the application when interacting with the human-like characters (Robert, Gabrielle, Arman, Olivia) than they do when the interaction is with Kai. While the overall rate of approval is the same for both types of persona, we find significantly more `Strongly agree' responses when users evaluate the former group. On the other hand, results are less conclusive when the chatbot is assessed for empathy. In this context, the human-like personas still receive more top range responses, however, when considering both `Agree' and `Strongly agree' answers, Kai is scored positively by a greater percentage of participants.

\subsection{Limitations of the study}

Of 23 non-clinician volunteers that signed up to the study only 16 completed the evaluation questionnaire, resulting in a further reduction of an already modest sample. Moreover, since the questionnaires are anonymous, we do not know how the sex and age distribution of our actual sample (i.e. those who returned a completed questionnaire) compares to that of the entire pool of volunteers. To design an effective future trial, this distribution should be considered carefully. For example, we have noted in this study that users favoured male characters over female ones (60\% of all interactions were with Robert and Arman). This may be due to the fact that males were over represented in our group of volunteers, and repeating the evaluation with a more evenly distributed sample could help validate or disprove this hypothesis. 

Moreover, increasing the number of required interactions and the length of the intervention in the future may give participants a more informed opinion of the strengths and weaknesses of the chatbot, as some of these (e.g. its ability/inability to present the user with novel utterances over time) may only be evident over a period longer than five days.

Finally, fluency and diversity are two main objectives of our conversational framework, yet we evaluate them only indirectly by asking participants how engaging they found the bot's conversations. We do this to be consistent with the evaluation data collected in the previous trial, and thus have a dependable baseline for comparing our results. However, this leaves us with little insight into why a minority of the participants have found the bot not to be engaging. In future studies, it would be advisable to have the chatbot's dialogue explicitly evaluated for fluency and diversity.

\subsection{Future work}

Despite obtaining encouraging results, the chatbot's emotion classifier has room for improvement. Four emotional contexts are hardly sufficient to cover an acceptable range of human emotions. Collecting more data relative to different contexts as well as more nuanced feelings would thus be necessary to train a more competent model. Moreover, as more than one emotion can be felt and expressed at the same time, the classification problem could be cast as a multi-label one \cite{Jabreel19}.

In addition, as highlighted by trial participants, individuals may not necessarily feel better or worse after completing SAT protocols. When asking for feedback, the bot should therefore accept as a valid answer the fact that a user might have detected no change in their mood.

Lastly, to increase the safety of the platform and its applicability in wider contexts, future implementations should include a mechanism to recognise and respond to any user input suggestive of self-harm or suicidal thoughts. Such a mechanism could consist of a keyword-based model for detecting terms commonly associated with self-harming~\cite{harvey2012}. Upon detection of any of these terms, the bot should promptly direct users to dedicated emotional support and/or suicide prevention services available in the country where they are located (this may require collection of IP addresses or other geolocation data).

\section*{Acknowledgment}

The authors would like to thank Noor Research (Atlanta, USA) for funding this work and the Empowered Human Foundation (Canada) for sponsoring the trial.

\bibliography{references.bib}
\bibliographystyle{IEEEtran}

\end{document}